\documentclass[a4paper,10pt,twoside]{article}

\usepackage{clin}        
\usepackage{harvard}     
\usepackage{hyperref}
\usepackage{booktabs}
\usepackage{tabularx} 
\usepackage{geometry}
\usepackage{array}
\usepackage{caption} 
\usepackage{makecell} 
\usepackage{pgfplots}
\usepackage{amsmath}
\usepackage{tikz}
\usepackage{multirow}
\usepackage{xcolor}
\usepackage{soul}
\definecolor{customcolor}{rgb}{0.8, 0.6, 0.0}
\usetikzlibrary{calc,patterns,decorations.pathmorphing,decorations,fit}
\pgfplotsset{compat=1.18}
\usepgfplotslibrary{external}

\pgfplotsset{compat=1.18}

\usepackage{tikz}
\usetikzlibrary{shapes.geometric, arrows, positioning}

\begin{document}
\raggedbottom
\title{Assessing Dutch Syllabification Algorithms and Improving Accuracy by Combining Phonetic and Orthographic Information through Deep Learning}

\author{Gus Lathouwers$^*$ \email{guslathouwers@gmail.com}\\
{\normalsize \bf Wieke Harmsen}$^*$ \email{wieke.harmsen@ru.nl}\\
{\normalsize \bf Catia Cucchiarini}$^*$ \email{catia.cucchiarini@ru.nl}\\
{\normalsize \bf Helmer Strik}$^*$ \email{helmer.strik@ru.nl}\\
\AND \addr{$^*$Radboud University, the Netherlands}}

\maketitle\thispagestyle{empty} 


\begin{abstract}
Syllabification describes the task of dividing words into syllables. Due to many rules and exceptions, training an algorithm to perform syllabification with high accuracy remains a challenge. Throughout the last decades, different algorithms have been put forth for Dutch syllabification, yet a comprehensive comparative assessment has not been done. Additionally, deep learning has gained significant popularity within NLP in recent years, yet no modern deep-learning based framework has been developed for Dutch orthographic syllabification. Finally, phonetic and orthographic syllabification algorithms have been examined separately, but not in combination. The aim of the current research was twofold: (a) to examine the performance of existing Dutch syllabification algorithms, and (b) to investigate whether combining phonetic and orthographic information into a single model can increase syllabification performance. To compare the performance of algorithms, four algorithms (Brandt Corstius, Liang, Trogkanis-Elkan (CRF), and a newly conceived deep-learning model) were applied to three different datasets (dictionary words, loanwords, pseudowords). The algorithms show varying performance across datasets, with the data-driven algorithms outperforming a knowledge-based algorithm in all but one condition. The new deep-learning methods developed led to increased performance compared to the best found in the literature (99.65\% word accuracy, a 0.14\% improvement). An analysis of the words for which adding phonetic information improved syllabification performance indicates that these were words in which the orthographic ambiguity could be resolved by information on pronunciation. Future research could examine other areas where phonetic information can benefit orthographic processing. In addition, the newly developed deep learning frameworks can be applied to other languages than Dutch.

\end{abstract}

\section{Introduction}

Syllabification describes the task of dividing words into syllables. Automated syllabification by algorithms remains a complex challenge due to the many rules and exceptions that govern syllable division. For instance, Webster’s Dictionary lists more than twenty rules for dividing words into syllables in English, each with many idiosyncrasies and exceptions for individual words \cite{Gove1993}. For Dutch, syllable boundaries may be determined by different, sometimes competing, principles, such as the sonority principle, the maximum onset principle, and priority rules for prefixes and suffixes \cite{Schiller1996,Corstius1970}.

Beyond being of interest because of its theoretical complexity, syllabification plays an important role in underlying several Natural Language Processing (NLP) technologies, such as text-to-speech \cite{Pradhan2013,Sarma2020} and grammar processing applications \cite{Hauer2013,Dascalu2017}. Syllable division can also be used to select texts with syllable properties, such as syllable length or specific syllable structures \cite{Alfiansyah2018}. Syllable analysis has also been used in document analysis to provide a benchmark for text complexity \cite{Ayyaswami2019,Man2021} or as a tool for rhyme analysis through identifying the last syllable of words \cite{Marco2021}.

In spite of its relevance, research on Dutch syllabification algorithms is outdated, with the last contribution stemming from \citeasnoun{Trogkanis2010}. Existing research has generally been focused on developing new methods, instead of benchmarking existing ones. Novel deep-learning methods which have enjoyed widespread popularity within NLP in the last decade\footnote{See \citeasnoun{Ahmed2023} for an overview.} have been used to create new deep learning solutions for syllabification in some languages \cite{Corlatescu2022}, yet not the Dutch language. One advantage of deep-learning approaches is that they allow the incorporation of different sources of information, such as phonetic and orthographic data \cite{Gale2023}, which could ostensibly help improve syllabification accuracy.

In this paper we present research that assesses existing Dutch syllabification algorithms by applying them to different datasets. In addition, we investigate whether combining both phonetic and orthographic information through the use of novel deep-learning techniques leads to better syllabification performance.

\subsection{Dutch syllabification algorithms}

In this section, we present a brief overview of algorithms developed for syllabification over the years. Initial work by \citeasnoun{Corstius1970} employed a knowledge-based method to syllabify words, meaning it syllabified according to explicitly outlined grammar rules manually coded by Brandt Corstius. Two decades later, \citeasnoun{Daelemans1989} expanded on Brandt Corstius’ work by incorporating a lexical library among other improvements, allowing for more consistent syllabification results. Simultaneously, challenges related to ambiguities from sentence context and semantic variation spurred a shift towards probabilistic and machine learning approaches. Specifically, \citeasnoun{Liang1983} introduced a different pattern-based approach to hyphenating words, which gained popularity because it was language-agnostic, meaning it was not tied to language-specific spelling rules and could thus be applied to any language. Work by \citeasnoun{Boot1984} further explored pattern-based processing and its application to Dutch.

One issue that is important to clarify is the use of the terms ”hyphenation” versus ”syllabification.” Although these terms are sometimes used interchangeably in the literature, they refer to inherently different ways of processing, with hyphenation describing word breaking according to spelling and grammar principles (for Dutch, the \textit{Groene Boekje} rules)\cite{Tutelaers1993}, whereas syllabification follows pronunciation conventions. For the purposes of text analysis in language processing, syllabification is often preferred, as it matches syllable division in spoken language, which in turn can be used for applications that rely on the phonetic structure of words. See Table~\ref{tab:differences} for examples of words where hyphenation and syllabification produce different division patterns. 

While authors such as Brandt Corstius, Liang, and Boot focused their efforts on more traditional pattern- or knowledge-based systems initially, developments afterward saw interest in novel lines of machine learning technologies. The first instance of a neural network applied to Dutch was formulated by \citeasnoun{Daelemans1992}, who used a backpropagation setup to syllabify texts. Afterwards, \citeasnoun{Bouma2003} employed a different finite-state method for syllable boundary prediction that relies on automatic rule abstraction. Approaching the problem from yet another machine learning perspective, \citeasnoun{Bartlett2008} combined sequence prediction with support vector machines, showing good results across multiple languages, including Dutch and English. The most recent method applied to Dutch syllabification was developed by \citeasnoun{Trogkanis2010}, who used a Conditional Random Field (CRF)-based approach for syllabifying Dutch and English, achieving comparable results to \citeasnoun{Bartlett2008}.

\subsection{Comparison of syllabification algorithms}

Due to the large number of new machine learning methods being developed within NLP in the last decades, the importance of comparing existing algorithms has been noted, rather than just introducing new ones \cite{Strobl2022}. Pitfalls of introducing new models without testing against existing ones include over-optimization on customized datasets, as well as selective use of metrics that may present an algorithm in a more favorable way \cite{Niessl2021}. For these reasons, establishing a performance baseline across algorithms \cite{Boulesteix2013,Friedrich2023} and creating a database of multiple datasets for future testing \cite{Yousefi2009b} could greatly improve understanding of algorithm performance tendencies. Such systems may help reduce reporting bias and ensure consistent testing practices on carefully vetted datasets \cite{Phang2021}.

\begin{table}[t]
\captionsetup{justification=raggedright, singlelinecheck=false} 
\centering
\begin{tabular}{ l | l l | l }
\toprule
\textbf{Full Dutch Word} & \textbf{Hyphenated\footnotemark} & \textbf{Syllabified} & \textbf{Translation} \\ \midrule
eland         & eland                & e-land         & moose               \\
atoomenergie  & atoom-ener-gie        & a-toom-e-ner-gie & atomic energy       \\
gloria        & glo-ria              & glo-ri-a        & gloria              \\
aliënatie     & ali-ena-tie          & a-li-e-na-tie   & alienation          \\
bakoven     & bak-oven          & bak-o-ven   & baking oven   
\\
bioscoop     & bio-scoop          & bi-o-scoop   & cinema 
\\
ruïne     & ru-ine          & ru-i-ne   & ruin 
\\\bottomrule
\end{tabular}
\caption{Examples of Dutch words with different syllabification and hyphenation word divisions.}\label{tab:differences}
\end{table}

\footnotetext{Retrieved from \url{https://github.com/jitsi/jiwer}}

In the context of Dutch syllabification, no standardized review of existing algorithms exists, yet there are reasons to believe such a comparison could be of value. Previous research on Dutch syllabification reports varying accuracy levels \cite{Bouma2003,Bartlett2008,Trogkanis2010}, but use different, sometimes incompatible metrics. For example, hyphen accuracy (the number of hyphens in words correctly predicted) is difficult to reconcile with character accuracy (the number of correctly predicted characters followed by a hyphen) or word accuracy (the number of words with all syllables predicted correctly) \cite{Bouma2003,Adsett2009,Trogkanis2010}. Differences in dataset filtering across studies also result in datasets with varying word makeup and size \cite{Bartlett2008,Trogkanis2010}. A comparison using a single standardized dataset could help provide a coherent view of algorithm performance.

\subsection{Deep learning methods}

Another area that has seen increased interest in the last decade concerns deep learning methods. Beyond providing improved performance on many NLP tasks, deep learning may especially excel in complex pattern analysis by utilizing neural-based pattern recognition techniques \cite{Ahmed2023}. For syllabification, neural based methods have been developed as well (for Dutch, starting with Daelemans and van den Bosch, 1992), showing potential for increased performance. Recent lines of deep learning models typically employ techniques such as convolutional neural networks (CNNs) and Bidirectional Long Short-Term Memory (BiLSTM), which have proven effective in tasks such as sequence modeling and feature extraction, due to their capacity for high-level pattern analysis \cite{Krantz2019,Corlatescu2022}.

One key advantage of deep learning approaches is the degree of customizability in model design \cite{Sarker2021}. In many cases, deep learning can be applied in an unsupervised manner, where models may autonomously learn to recognize patterns from raw data without the need for manually specified features \cite{Ahmed2023}. This allows for multi-model layering setups, such as those used in data augmentation \cite{Gavrishchaka2018}, or integrating multiple sources of information into a single model \cite{Gale2023}.

In addition to orthographic syllabification, a secondary area of interest has been applying syllabification algorithms to phonetic representations of words. Syllabification algorithms have traditionally been applied to orthographic \cite{Trogkanis2010} and phonetic word representations \cite{Krantz2019} in isolation, but the two have not been combined yet. Recent syllabification research has moved toward integrating orthographic with phonetic input by analyzing linguistic similarities between orthographic and phonetic words \cite{Tits2023}. To this effect, deep learning mechanisms such as attention may provide a means of creating overarching models that combine the two.

\subsection{Current study}

The current study addresses the following two research questions: 
\begin{enumerate}
    \item \textbf{RQ1: What is the comparative performance of different algorithms tested under identical conditions?} \\
    To answer this question, we first provide a comparative overview of existing syllabification algorithms. Previous research is fragmented with regard to methodology and datasets used, making it difficult to compare performance and generalizability across datasets. Following this, several existing Dutch syllabification algorithms will be tested under identical conditions. Additionally, a newly developed deep learning framework will be applied to Dutch orthography. All algorithms will be tested on multiple datasets that have been curated \textit{a priori} to allow for generalization benchmarking.
    \item \textbf{RQ2: How does the addition of phonetic information affect Dutch orthographic syllabification performance?} \\
    Existing research has investigated the performance of orthographic \cite{Trogkanis2010} and phonetic \cite{Krantz2019} syllabification separately, but not in combination. Following an analysis of the synergy between these two information sources \cite{Tits2023}, this study explores the potential of combining phonetic and orthographic information into a single model.
\end{enumerate}

\section{Methodology}

To answer the two research questions, two experiments were designed. Experiment 1 focuses on a comparative assessment of Dutch syllabification algorithms. Experiment 2 examines the relationship between orthographic and phonetic information and their combined effect on syllabification performance through a combined deep learning model.

\subsection{Datasets}

To evaluate the performance of the algorithms, three different datasets were compiled or retrieved from various sources\footnote{All datasets and algorithms used in this study can be accessed at \url{https://github.com/guslatho/syllabify-torch}}. Previous research \cite{Bouma2003,Trogkanis2010,Bartlett2008} exclusively relied on Dutch standard dictionary words to assess algorithm performance. In addition to testing performance on dictionary words, the current study also includes two other datasets: a set of loanwords and a set of Dutch pseudowords. Table~\ref{tab:datasetsused} provides an overview of the datasets used and their characteristics.

\textbf{CELEX.} The CELEX (Dutch Centre for Lexical Information) database was used to measure algorithm performance on standard Dutch dictionary words. To allow for direct comparison with earlier research, the final list used was the same as that employed in the study by \citeasnoun{Trogkanis2010}, who distributed the list online \footnote{\url{https://cseweb.ucsd.edu/~elkan/hyphenation/}}. The total number of words included is n=293{,}747, comprising words typically found within the Dutch language, such as verbs, nouns, adverbs, and various other parts of speech (see Appendix~\ref{appendix:samplewords} for a sample). In their paper, Trogkanis and Elkan describes filtering out words that had multiple syllabification solutions (for instance, \textit{zoeven} can be syllabicated to either \textit{zoe-ven} or \textit{zo-e-ven} depending on its meaning). Manual inspection of the list showed that these were still present in the dataset they distributed online, thus these were removed (n=33), resulting in a list of n=293{,}714.

\textbf{Loanwords.} A set of loanwords was included to test the performance of algorithms on words common in the Dutch language, but originating from other languages, such as English and French. This allowed testing the adaptability of different algorithms to more challenging word conditions. A list of 10{,}372 words was retrieved from \citeasnoun{VanDerSijs2005}. Of these ten thousand words, n=8{,}015 were found to be already present in the original CELEX word list used in the first dataset and were thus removed. To exclude dated or archaic words from the list, the remaining words were cross-checked with Gigant-Molex, a library containing contemporary Dutch words. The final resulting dataset consisted of n=1{,}135 words. Appendix~\ref{appendix:samplewords} lists a sample of the words included.

\textbf{Pseudowords.} Pseudowords refer to non-existent words in the Dutch language that are artificially created to resemble Dutch word forms. A list of Dutch pseudowords found in the CHOREC (Children’s Oral Reading Corpus\footnote{\url{https://taalmaterialen.ivdnt.org/download/tstc-chorec-spraakcorpus}}) was used as a reference point. The aim of the pseudoword set was to measure the performance of algorithms on words that conform to Dutch word construction rules, but do not appear in the CELEX database. The CHOREC list contains 120 words, of which n=21 were found to have multiple syllabification solutions (e.g., for \textit{menuur}, both \textit{men-uur} and \textit{me-nuur} are grammatically correct syllable solutions in Dutch) and were removed. This resulted in a set of n=99 words. Appendix~\ref{appendix:samplewords} lists a subsection of the words included.

\begin{table}[t!]
\captionsetup{justification=raggedright, singlelinecheck=false} 
\centering

\begin{tabular*}{\textwidth}{@{\extracolsep{\fill}} l c c l p{0.4\textwidth} @{}}
\toprule
\textbf{Dataset} & \multicolumn{2}{c}{\textbf{Size (n)}} & \textbf{Origin} & \textbf{Description} \\ \cmidrule(r){2-3}
 & \textbf{Training} & \textbf{Testing} &  & \\ \midrule
Dictionary & 264{,}343 & 29{,}371 & CELEX & Common words found in the Dutch dictionary. The CELEX dataset was used as a training and testing set with a 90\%/10\% split. \\ 
Loanwords & - & 1{,}135 & De Sijs (2009) & Words found in the Dutch language originating from other languages, e.g., French or English. Used as a testing set only.\\ 
Pseudowords & - & 99 & CHOREC & Words artificially created to imitate Dutch word conventions. Used as a testing set only.\\ 
\bottomrule
\end{tabular*}
\caption{Datasets used for algorithm training and testing.}
\label{tab:datasetsused}
\end{table}

\subsection{Algorithms}

To compare the effectiveness of existing syllabification algorithms, the performance of three Dutch syllabification algorithms was analyzed. Two of these algorithms (Brandt Corstius, CRF) were replicated from the literature, while new weight patterns were generated for a third algorithm (Liang) adapted to Dutch syllabification. Additionally, a novel deep-learning method was developed.

\subsubsection{Brandt Corstius}
The knowledge-based algorithm proposed by \citeasnoun{Corstius1970} is notable for being a linguistic algorithm built specifically on Dutch spelling and grammar rules. It operates by first compressing words into different vowel and consonant clusters, for example marking common Dutch vowel combinations such as \textit{eu}, \textit{au}, \textit{ieu} as distinct entities. Words are subsequently processed by a combination of maximum onset principle and consonant-priority patterns. For example, \textit{loonbrief} (pay slip) is first compressed into [l][oo][n][b][r][ie][f], for which maximum onset processing for valid consonant combinations produces \textit{loon-brief}.

Brandt Corstius originally developed his algorithm in Algol-70 and tested it on a sample of 43{,}712 word tokens compiled from Dutch newspapers, poems, and books. He reported a word error rate of only 0.5\%. Brandt Corstius notes that his linguistic knowledge-based algorithm is effective for most Dutch words, but struggles with compound words where the maximum onset principle may falter. In the current study, a recreation was made in Python, following the logic of his system as outlined in his paper.

\subsubsection{Liang}
Originally introduced for hyphenation purposes in LaTeX, the algorithm by \citeasnoun{Liang1983} has since seen widespread usage in various software applications. Liang’s algorithm functions by storing hyphenation and exception patterns, resulting in a list with weights. Unlike the Brandt Corstius algorithm, Liang’s algorithm is data-driven and non-language specific, and thus can be applied to any dataset with hyphenation or syllabification solutions agnostically.

Various studies have assessed the performance of Liang’s algorithm. A comparison study by \citeasnoun{Adsett2009} found it to be effective at word hyphenation and syllabification, reaching a mean word accuracy of roughly 96\% tested across nine languages (Adsett and Merchand did not report accuracy for individual languages). For Dutch, a word accuracy rate of 93.08\% is reported \cite{Trogkanis2010}. However, the Trogkanis and Elkan study describes applying hyphenation rules in training the Liang algorithm on their dataset, which may result in a skewed representation of performance. For the current study, custom patterns were generated for Dutch that conform to syllabification conventions, using the original TeX software tool \texttt{PATGEN.EXE}, found at \url{https://www.tug.org/texlive/doc/texlive-en/texlive-en.html}.

\subsubsection{Conditional random field algorithm}
This approach employs Markov chain principles through a conditional random field implementation. Developed by \citeasnoun{Trogkanis2010}, the CRF model processes input words by capturing windows around each character for syllable prediction. For each input character a \texttt{'0'} or \texttt{'1'} is assigned, with \texttt{'0'} signifying no syllable break following the character, and \texttt{'1'} signifying a syllable break following. Trogkanis and Elkan report excellent word accuracy rates, with 99.51\% of the words in their dictionary words set syllabified correctly.

Trogkanis and Elkan’s model was originally implemented in CRF++, a software package for conditional random field modeling. The current implementation was developed in Python, using Chaine, an open-source package for creating CRF models. A few optimizations were found to benefit performance of the model. A 6-window size for each character was applied instead of a 5-character window, as described by Trogkanis and Elkan. Start-of-word and end-of-word tokens were also added to better capture word boundaries, and the coding scheme was slightly modified.

\subsubsection{Deep learning model (new model)}
A new deep-learning model was developed in TensorFlow and subsequently ported to Torch\footnote{Template for the Torch training model and weights for Dutch can be found at \url{https://github.com/guslatho/syllabify-torch}}. Previously, CNN, BiLSTM, and CRF components were shown to be effective for syllabification of phonemes \cite{Krantz2019} and Romanian syllabification \cite{Corlatescu2022}, and were thus chosen as the starting components for the architecture of the current model. However, the design of the current model differs by using a sequential setup, where individual letter-windows of a word are first locally processed through convolution layers before being combined at a higher-order BiLSTM/CRF layer.

The architecture of the deep learning model consists of three parts. First, a 5-character window is formed around each letter of the input word. This window-based approach allows the model to focus on local information around each character and detect nearby patterns. The 5-character window is first passed into an embedding layer, followed by a convolution layer, dropout, and global pooling layer.  The entire first block—embedding, convolution, dropout, and pooling—can be conceptualized as a ‘micro-level’ local analysis for extracting information around each character.

For each 5-character window for each letter analyzed this way during the convolution stage, the output is flattened and forwarded to a stack of two BiLSTM layers. The BiLSTM layers integrate the local 'micro-level' analyses information into larger patterns. Finally, the second BiLSTM outputs to a CRF, which makes predictions for each character of the input word. Here, '0' represents no syllable boundary following the letter, and '1' represents a syllable boundary. As in Trogkanis and Elkan's (2010) coding scheme, the binary string forms the prediction for the final output. For example, the Dutch word input \textit{berekening} (calculation), after being processed by the model, would result in the string '0101010000'. This binary string would denote hyphens after the 2nd, 4th, and 6th letters, resulting in the output syllabification \textit{be-re-ke-ning}.

Token length was set at 34, in line with the maximum word length found in the training wordset; words were post-padded. The addition of L2 normalization (batch norm) was found to increase model consistency in early testing and was therefore applied. The training batch size was set at 64. The initial embedding layer was set at 128 units, producing an input shape of \texttt{[64 x 34 x 5 x 128]} for \texttt{[batch x token x window x embedding]}. The convolution layer was set at 40 filters with a kernel size of 3 and a sliding window of 1. Following a dropout rate of 0.3, a global max pooling layer with a kernel of 3 was applied. The output was then flattened, resulting in a \texttt{[64 x 34 x 200]} shape, which was forwarded to the first BiLSTM layer. The unit size for both BiLSTM layers was set at 128, producing a shape of \texttt{[64 x 34 x 256]}, which was forwarded to the final CRF layer.

\subsection{Training and testing procedure}

Of the four algorithms used in the benchmark analysis, three required a training library: Liang, CRF, and the deep learning model. The fourth algorithm, Brandt Corstius, syllabified according to predefined Dutch language rules, and thus required no training data. For these three data-driven algorithms, 90 percent of the CELEX dataset was used for training and the remaining 10 percent for testing. Similar to the ten-fold cross validation approach used by Trogkanis and Elkan (2012), the deep learning network was trained on ten 90/10 splits generated by ten consecutive random seeds. The Liang and CRF libraries produced consistent results; as such, only five random 90/10 splits were sampled for each algorithm.

Whereas Chaine for CRF and \texttt{PATGEN.EXE} for Liang contained automatic stopping procedures, the deep learning model could be trained indefinitely. As such, a limit of 180 training epochs was set. Practice trials revealed small increments in training after 180 epochs, but due to time considerations, the limit was set at 180 epochs. Training time for the CRF algorithm was approximately 40 minutes per model. For the deep learning model, GPU acceleration resulted in a mean training time of roughly 2.5 hours using a modern GPU (RTX 4080).

\subsubsection{Evaluation metrics}

In total, six metrics were used to assess algorithm performance.

\textbf{Word/Character Accuracy Metrics.} Given syllabification is a sequential prediction task at the word level, algorithm performance is commonly measured using Overall Word-Level Error Rate (OWER\%) \cite{Adsett2009,Trogkanis2010,Bartlett2008}. However, other metrics such as Overall Letter-Level Error Rate (OLER\%) \cite{Trogkanis2010} and Hyphen Error Rate (HER\%) have also been used \cite{Bartlett2008,Bouma2003}. Given the widespread use of these different metrics, all three have been included here.

Overall Word-Level Error Rate expresses the proportion of words incorrectly syllabified by an algorithm (Equation \ref{eq:OWE}). It is usually expressed as a percentage, with an OWER\% of 10 indicating that one out of ten words in the test set were predicted incorrectly in its entirety. Overall Letter-Level Error Rate describes the total number of individual characters across the full word set for which a prediction ('0' for no syllable follows, '1' for a syllable follows) was made incorrectly (Equation \ref{eq:OLER}). Since most letters in a word are not followed by a syllable, OLER\% is susceptible to be inflated by a high number of true negatives. Lastly, Hyphen Error Rate (HER\%) represents the number of hyphens not predicted out of all hyphens (Equation \ref{eq:HER}). Like OLER\%, HER is liable to distort algorithm performance since it does not penalize false hyphen predictions.\footnote{It should be noted that, in the current study, HER\% refers to the error rate in relation to syllabification, not hyphenation. Given its widespread use in syllabification research, the same term will be used here out of convention, even though it is technically incorrect since hyphenation refers to word breaking according to spelling and grammar principles, not pronunciation.}

\begin{equation}
OWER\ \% = \frac{Words\ containing\ at\ least\ one\ FP\ or\ FN } {Total\ words}
\label{eq:OWE}
\end{equation}

\begin{equation}
OLER\ \% = \frac{FN + FP}{FN + TP + TN + FP}
\label{eq:OLER}
\end{equation}

\begin{equation}
HER\ \% = \frac{FN}{TP + FN}
\label{eq:HER}
\end{equation}

As an example for application of these metrics, for the Dutch word 'we-reld-be-ker' (\textit{world cup}), an incorrect solution of 'wereld-beker' would result in an Overall Letter-Level Error Rate of 18.18\%, since 2 out of 11 characters were predicted incorrectly. Hyphen Error Rate would be 66.67\%, with only 1 of 3 syllable boundaries predicted correctly. Overall Word-Level Error Rate would classify the whole word as incorrectly syllabified due to the presence of two false negatives.

\textbf{Precision, recall, F1.} In addition to the three metrics above, Precision (Equation \ref{eq:Precision}), Recall (Equation \ref{eq:Recall}), and F1 (Equation \ref{eq:F1}) were also included to assess algorithm performance. Note that Recall is the complement of HER, expressed as \( R = 100\% - \text{HER} \).

\begin{equation}
\text{Precision} = \frac{TP}{TP + FP}
\label{eq:Precision}
\end{equation}

\begin{equation}
\text{Recall} = \frac{TP}{TP + FN} = 100 - HER
\label{eq:Recall}
\end{equation}

\begin{equation}
F_1 = \frac{2 \cdot \text{Precision} \cdot \text{Recall}}{\text{Precision} + \text{Recall}}
\label{eq:F1}
\end{equation}

\section{Results}

\subsection{Experiment 1: comparison of algorithms on Dutch datasets}

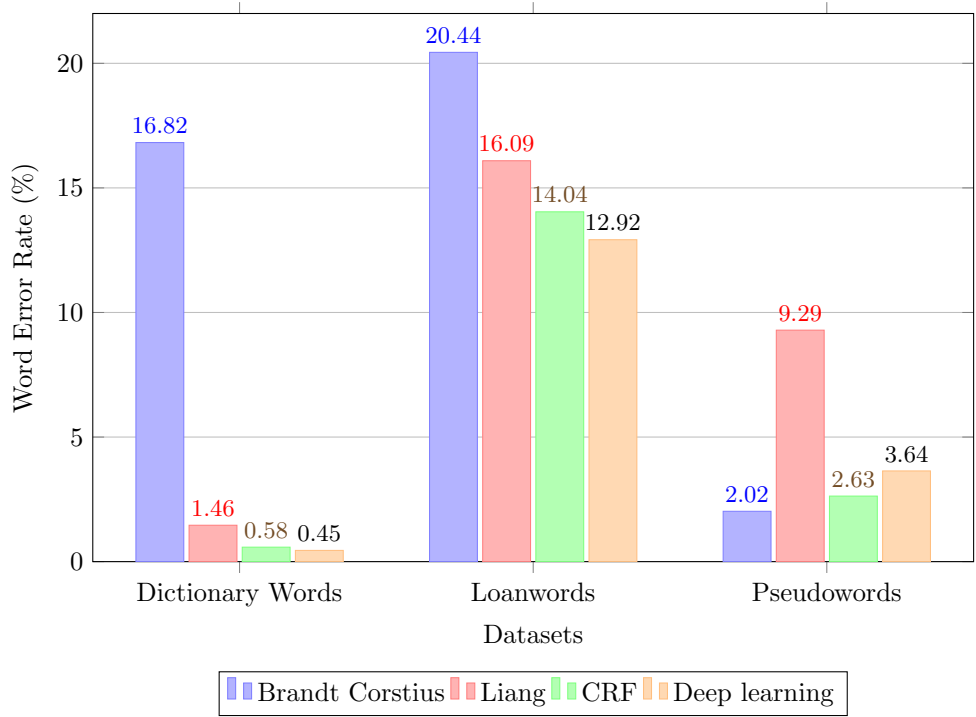
\begin{figure}[t]
    \centering

\begin{tikzpicture}
\begin{axis}[
    title={ },
    ylabel={Word Error Rate (\%)},
    xlabel={Datasets},    
    symbolic x coords={Dictionary Words, Loanwords, Pseudowords}, 
    xtick=data,
    ymin=0, ymax=22,
    ymajorgrids=true,
    bar width=18pt, 
    width=0.9\textwidth,
    height=0.6\textwidth,
    legend style={at={(0.5,-0.2)}, anchor=north, legend columns=-1},
    legend cell align={left},
    enlarge x limits=0.25, 
    nodes near coords,
    every node near coord/.append style={font=\small},
    ybar, 
]

\addplot+[
    draw=blue!50,
    fill=blue!30,
] coordinates {
    (Dictionary Words, 16.82) (Loanwords, 20.44) (Pseudowords, 2.02)
};
\addlegendentry{Brandt Corstius}

\addplot+[
    draw=red!50,
    fill=red!30,
] coordinates {
    (Dictionary Words, 1.46) (Loanwords, 16.09) (Pseudowords, 9.29)
};
\addlegendentry{Liang}

\addplot+[
    draw=green!50,
    fill=green!30,
] coordinates {
    (Dictionary Words, 0.58) (Loanwords, 14.04) (Pseudowords, 2.63)
};
\addlegendentry{CRF}

\addplot+[
    draw=orange!50,
    fill=orange!30,
] coordinates {
    (Dictionary Words, 0.45) (Loanwords, 12.92) (Pseudowords, 3.64)
};
\addlegendentry{Deep learning}

\end{axis}
\end{tikzpicture}
\caption{Word accuracy rates (OWER\%) for algorithms across datasets.}
\label{fig:worderrorrates}

\end{figure}

Four algorithms, Brandt Corstius, Liang, CRF, and Deep Learning (DL), were tested against three dataset. See Figure~\ref{fig:worderrorrates} for a bar plot containing the word accuracy rates per datasets. A complete overview of metrics including standard deviations is listed in Table~\ref{tab:comparisonresults}.

\subsubsection{Results}

For the first dataset, dictionary words (n=29{,}375), the deep learning model showed the best overall performance (OWER\%, 0.446; OLER\%, 0.068; HER\%, 0.145; Precision, 99.852; Recall, 99.855; F1, 99.853). The Overall Word-Level Error Rate of 0.446 indicates less than 1 in 200 words being syllabified incorrectly in the test set. The second-best performing algorithm was the CRF model, showing a 0.129 higher word error rate than the deep learning model (OWER\%, 0.575; OLER\%, 0.087; HER\%, 0.178; Precision, 99.807; Recall, 99.822; F1, 99.814). The Liang and Brandt Corstius algorithms scored worse than CRF and the deep learning model across all metrics. For Overall Word-Level Error Rate, Liang’s algorithm had a rate of 1.460\%, whereas Brandt Corstius scored above the 10\% threshold for incorrectly syllabified words (Overall Word-Level Error Rate of 16.819\%, translating to 83.181\% of all dictionary words syllabified correctly).

For the loanwords dataset (n=1{,}135), all algorithms scored worse in comparison with their performance on the dictionary words set. The deep learning algorithm showed superior performance on all metrics except precision (OWER\%, 12.916; OLER\%, 2.433; HER\%, 4.666; Precision, 93.870; Recall, 95.334; F1, 94.596). The Brandt Corstius algorithm showed the worst performance across all metrics (with an OWER\% of 20.441 meaning less than 4 out of 5 loanwords were syllabified correctly). Liang's algorithm scored better than the Brandt Corstius algorithm, but worse than CRF and the deep learning algorithm (OWER\%, 16.088; OLER\%, 2.571; HER\%, 6.576; Precision, 95.258; Recall, 93.424; F1, 94.332).

On the last word set, pseudowords (n=99), the performance of the algorithms followed a different trend in comparison to the first two datasets. Whereas Brandt Corstius was least accurate when applied to the first two datasets (dictionary and loanwords), in the pseudowords condition it showed the lowest Overall Word-Level Error Rate (OWER\% of 2.02). CRF showed a similar performance, scoring lower on Overall Word-Level Error Rate but slightly higher on character-level metrics in comparison with Brandt Corstius (OWER\%, 2.626; OLER\%, 0.395; HER\%, 0; Precision, 97.418; Recall, 100.000; F1, 98.691). Liang's algorithm performed comparatively worse than any other algorithm (OWER\%, 9.293; OLER\%, 2.094; HER\%, 4.694; Precision, 91.041; Recall, 95.306; F1, 93.123).
\vfill 

\begin{table}[t]
\centering
\scriptsize
\begin{tabular}{|l|l||l|l|l|l|}
\hline
\textbf{Dataset} & \textbf{Metric} & \textbf{Deep Learning} & \textbf{CRF} & \textbf{Liang} & \textbf{Brandt Corstius*} \\
\hline
\multirow{6}{*}{CELEX} 
& OWER\% & \textbf{0.446 (±0.038)} & 0.575 (±0.037) & 1.460 (±0.073) & 16.819 (±0.125) \\
& OLER\%  & \textbf{0.068 (±0.007)} & 0.087 (±0.005) & 0.180 (±0.006) & 3.143 (±0.014) \\
& HER\% & \textbf{0.145 (±0.017)} & 0.178 (±0.016) & 0.438 (±0.011) & 6.818 (±0.023) \\
\cline{2-6}
& Precision   & \textbf{99.852 (±0.016)} & 99.807 (±0.011) & 99.669 (±0.013) & 93.350 (±0.037) \\
& Recall   & \textbf{99.855 (±0.017)} & 99.822 (±0.016) & 99.562 (±0.011) & 93.182 (±0.023) \\
& F1   & \textbf{99.853 (±0.015)} & 99.814 (±0.012) & 99.616 (±0.012) & 93.266 (±0.027) \\
\hline
\hline
\multirow{6}{*}{Loanwords} 
& OWER\% & \textbf{12.916 (±0.638)} & 14.044 (±0.16) & 16.088 (±0.321) & 20.441 \\
& OLER\%  & \textbf{2.433 (±0.116)} & 2.563 (±0.028) & 2.571 (±0.041) & 4.160 \\
& HER\% & \textbf{4.666 (±0.289)} & 6.840 (±0.079) & 6.576 (±0.072) & 7.815 \\
\cline{2-6}
& Precision   & 93.870 (±0.297) & \textbf{95.538 (±0.051)} & 95.258 (±0.211) & 89.898 \\
& Recall   & \textbf{95.334 (±0.289)} & 93.160 (±0.079) & 93.424 (±0.072) & 92.185  \\
& F1   & \textbf{94.596 (±0.256)} & 94.334 (±0.061) & 94.332 (±0.087) & 91.027  \\
\hline
\hline
\multirow{6}{*}{Pseudowords} 
& OWER\% & 3.636 (±1.794) & 2.626 (±0.553) & 9.293 (±1.317) & \textbf{2.020 } \\
& OLER\%  & 0.835 (±0.454) & \textbf{0.395 (±0.083)} & 2.094 (±0.271) & 0.455  \\
& HER\% & 2.041 (±1.273) & \textbf{0.000 (±0.000)} & 4.694 (±0.559) & 1.020  \\
\cline{2-6}
& Precision   & 96.492 (±1.759) & 97.418 (±0.531) & 91.041 (±1.209) & \textbf{97.980}  \\
& Recall   & 97.959 (±1.273) & \textbf{100.000 (±0.000)} & 95.306 (±0.559) & 98.980  \\
& F1   & 97.219 (±1.503) & \textbf{98.691 (±0.272)} & 93.123 (±0.869) & 98.477  \\
\hline
\end{tabular}

\caption{Accuracy means and standard deviations for metrics for algorithms tested across three datasets. Best performing algorithm for each metric is displayed in bold.}
\label{tab:comparisonresults}

\begin{minipage}{14.6cm}

\vspace{0.1cm}

\vspace{0.1cm}

\small(* The Brandt Corstius algorithm was only applied to the pseudowords and loanwords set once due to requiring no training library, as such no standard deviations is listed for these two sets.)

\end{minipage}

\end{table}

\subsubsection{Discussion}

The results of the first experiment show that the algorithms achieved high accuracy (fewer than 1 in 160 errors for the CRF and deep learning algorithms) when syllabifying dictionary words, which is consistent with findings in the literature \cite{Trogkanis2010,Bartlett2008}. All algorithms struggled more in the loanwords condition, which is expected since loanwords from other languages (e.g., English, French) often have syllable structures that differ from those of native Dutch words. 

One notable finding was that the knowledge-based algorithm by \citeasnoun{Corstius1970} showed the worst performance on dictionary words and loanwords, but performed best at the word level when syllabifying pseudo-words. This may initially seem counterintuitive, given that knowledge-based algorithms are explicitly designed to capture patterns in existing Dutch words and are not developed to generalize to unknown words. However, qualitative analysis shows that the Brandt Corstius algorithm is adept at syllabifying Dutch pseudowords that follow strict Dutch language conventions. The data-driven models, by contrast, struggled in certain scenarios because they overfit to patterns in the training set. For example, the pseudowords ‘jien’ and ‘leum’ were incorrectly syllabified as ‘ji-en’ and ‘le-um’ because the training set contained reference words with similar syllable breaks (e.g., ‘o-le-um’, ‘ju-bi-le-um’, ‘boei-en’, ‘ski-en’). The linguistic rules embedded in the Brandt Corstius algorithm, however, allowed it to recognize that the middle vowel patterns ‘ie’ and ‘eu’ should not be split, resulting in the correct classification of these as one-syllable words.

In comparison to the best results in the literature for Dutch, the current deep learning solution shows slightly increased performance on the same dataset (0.45\% word error rate instead of 0.49\% as reported by Trogkans and Elkan 2010). Furthermore, the Liang algorithm tested here shows a word error rate of only 1.46\% versus the nearly 7\% reported in the \citeasnoun{Trogkanis2010} study. A possible explanation for this discrepancy is that the current implementation was explicitly aimed at syllable-division instead of hyphenation. The results of the CRF implementation of the current study show slightly decreased performance to the CRF results mentioned by \citeasnoun{Trogkanis2010} (Trogkanis \& Elkan, 0.49 OWER\%; current study, 0.58 OWER\%), possibly owing to different software tools being used to implement the algorithm. Of interest are the results for F1, Recall, and Precision, which show the deep learning model achieving high balance between Precision and Recall, contributing to its high overall performance in the loanwords and CELEX datasets.

\subsection{Experiment 2: combining phonetic and orthographic information}

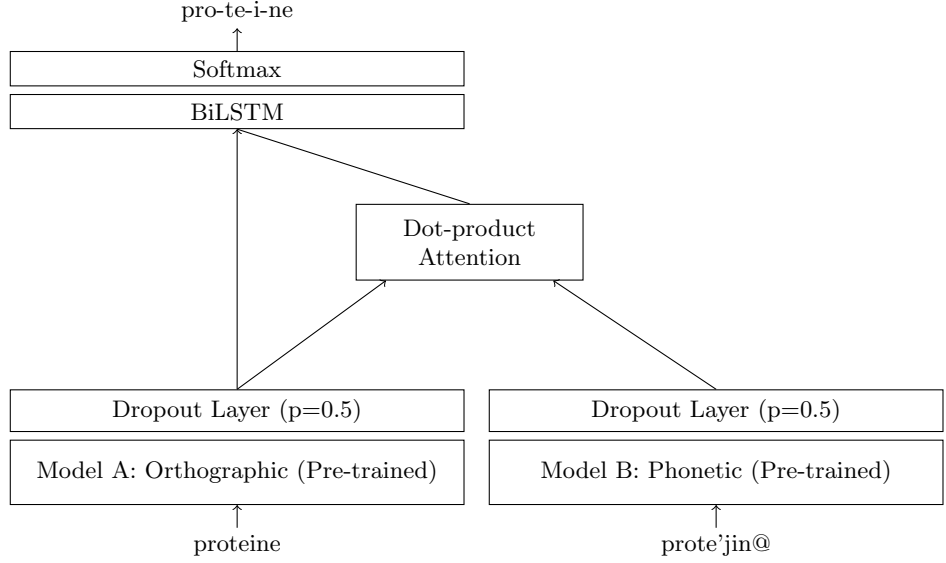
\begin{figure}[!t]
    \centering
    
\begin{tikzpicture}[scale=2,font=\small]

    
    \node [draw=black,minimum width=6cm,minimum height=0.85cm] (modelA) {Model A: Orthographic (Pre-trained)};
    \node [draw=black,minimum width=6cm,minimum height=0.85cm, right =0.32cm of modelA] (modelB) {Model B: Phonetic (Pre-trained)};

    \node[anchor=north, shift={(0,-0.7cm)}] at (modelA) {proteine};
    \node[anchor=north, shift={(0,-0.7cm)}] at (modelB) {prote'jin@};    

    \node [draw=black,minimum width=6cm,minimum height=0.2cm, above =0.1cm of modelA] (dropA) {Dropout Layer (p=0.5)};
    \node [draw=black,minimum width=6cm,minimum height=0.2cm, above =0.1cm of modelB] (dropB) {Dropout Layer (p=0.5)};

    \node [draw=black,minimum width=3cm,minimum height=1.0cm, above right=2.1cm and -1.44cm of modelA, align=center] (atten) {Dot-product \\ Attention};

    \node [draw=black,minimum width=6cm,minimum height=0.2cm, above =4.1cm of modelA] (bilstm) {BiLSTM};
    \node [draw=black,minimum width=6cm,minimum height=0.2cm, above =0.1cm of bilstm] (softmax) {Softmax};

    \node[anchor=south, shift={(0,0.5cm)}] at (softmax) {pro-te-i-ne};


    \draw[->] (dropA.north) -- ($(atten.south west) + (0.2cm, 0)$);
    \draw[->] (dropB.north) -- ($(atten.south east) - (0.2cm, 0)$);
    \draw[->] (dropA.north) -- (bilstm.south);
    \draw[-] (atten.north) -- (bilstm.south);    

    \draw[->] (softmax.north) -- ++(0,0.15cm);  
    
    \draw[->] ($(modelA.south) - (0, 0.15cm)$) -- (modelA);
    \draw[->] ($(modelB.south) - (0, 0.15cm)$) -- (modelB);

\end{tikzpicture}

\caption{Architecture of model C, combining model A and B.}
\label{fig:model_architecture}
\end{figure}

Given that the deep learning model was effective at Dutch syllabification, follow-up research was conducted to combine phonetic and orthographic word information. Using phonetic information as input requires access to phonetic word representations; fortunately, CELEX provides these for each word (for example, CELEX lists 'prote’jin@' for the orthographic notation 'proteine'). The goal was to investigate whether adding phonetic information improved orthographic syllabification accuracy and, if so, how this improvement was achieved. 

Since phonetic notations were not available for the loanwords and pseudowords datasets, these were excluded from the current analysis.\footnote{{During development, an early version incorporated a grapheme-to-phoneme component that generated phonetic representations for the model. While these setups performed well, the multi-stage nature of the pipeline led to a decision to exclude them here for reasons of parsimony. Future research could explore these and related approaches, such as automatic phoneme prediction during model training.}}

\subsubsection{Architecture}

In building a model that combines orthographic with phonetic input, a number of setups were experimented with, including transformer-based models. Ultimately, the model that appeared to be most effective was one that relied on combining pre-trained models. First, two instances of the deep learning model were trained on the orthographic and phonetic word presentations separately. Crucially, the orthographic model (model \textit{A}) and phonetic model (model \textit{B}) shared the same wordset, the only difference being that they were exposed to orthographic versus phonetic notations of the words.

After pre-training, both \textit{A} and \textit{B} were frozen and had their final BiLSTM and CRF layer removed. A new model, \textit{C}, was formed on top of the original two, by adding a BiLSTM layer and a new classification layer (softmax), see Figure \ref{fig:model_architecture} for an overview of the model. The orthographic output (\textit{A}) connected directly with the new BiLSTM layer, with dot-attention tying the output from \textit{B} to \textit{A}. The attention mechanism was important, given that the orthographic and the phonetic notations of words often do not line up automatically (for example, the orthographic notation ”proteine” is 8-length, while the phonetic notation of the same word ”prote’jin@” is 10-length). As a last addition, adding a high dropout layer (p=0.5) after \textit{A} and \textit{B} was found to be very effective in increasing performance. This is likely due to model \textit{C} not being stimulated to rely on new information  from \textit{B} in the absence of dropout layers. 

\subsubsection{Results}

\begin{figure}[t]
    \centering
    \hspace{-0.5cm} 
    \begin{tikzpicture}
        \begin{axis}[
            width=14.6cm, 
            height=8cm, 
            xlabel={Epoch}, 
            ylabel={Word Error Rate (\%)}, 
            axis y line=left, 
            axis x line=bottom, 
            minor tick num=1, 
            legend style={at={(0.5,-0.17)}, anchor=north}, 
            ymin=0.2, 
            ymax=0.8,  
            xmin=0,
            xmax=180
        ]
        
        \addplot[blue, thick] table[x=x, y=Graph1, col sep=comma] {data.csv};
        \addlegendentry{Orthographic Model};

        \addplot[red, thick] table[x=x, y=Graph2, col sep=comma] {data.csv};
        \addlegendentry{Phonetic Model};

        \addplot[green, thick] table[x=x, y=Graph3, col sep=comma] {data.csv};
        \addlegendentry{Combined Model};

        \end{axis}
    \end{tikzpicture}
    \caption{Word-level error rate after each epoch for the orthographic (blue), phonetic (red), and combined (green) models. The combined model, which integrates information from both the pretrained orthographic and phonetic models, shows a lower error rate on the validation set than the individual models.}
    \label{fig:validationerrorcomb}
\end{figure}
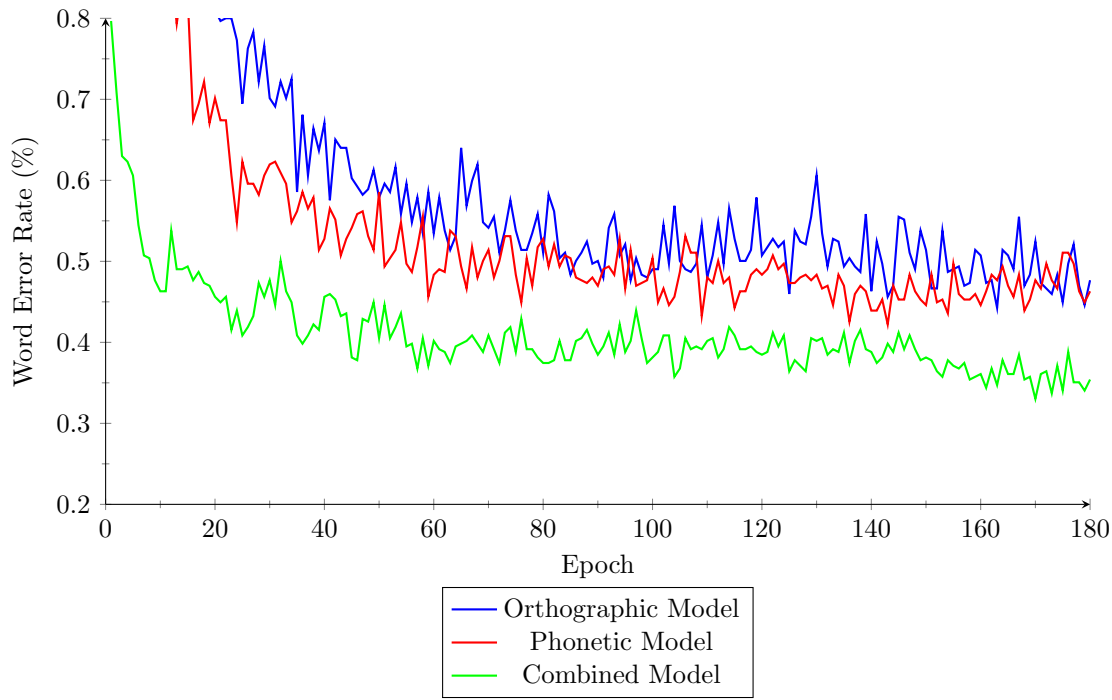

Figure \ref{fig:validationerrorcomb} displays the results for combined model \textit{C} as well \textit{A} and \textit{B}, the models it was built on, charted across training duration. The individual models both scored a respective 0.46\% (phonetic model) and 0.45\% (orthographic model) Overall Word-Level Error Rate. The combined model, \textit{C}, showed a decreased Overall Word-Level Error Rate, namely 0.35\% (OLER\% of 0.05,  HER\% of 0.12). See Appendix~\ref{appendix:improvedwords} for a sample of words that the combined model \textit{C} predicted correctly that the individual orthographic model failed. 

\subsubsection{Discussion}

In the second experiment, the inclusion of phonetic information through a deep-learning model was found to enhance syllabification performance. Combining phonetic information with orthographic information resulted in a word accuracy rate of 99.65\%, an absolute 0.14\% improvement over the best reported in the literature (Trogkanis \& Elkan, 2010; 99.51\%). Analysis of the words improved by the combined model shows that phonetic information can be particularly useful in providing additional context when orthographic information alone is ambiguous. For example, in the case of \textit{leptosoom}, the orthographic model incorrectly produced the output \textit{lep-tos-oom}, misattributing the first \textit{o} as a short vowel. In the phonetic representation, short and long vowels are coded differently, enabling the model to distinguish between these two vowel patterns and correctly produce \textit{lep-to-soom}.

\section{Discussion and Conclusions}

The goal of the current research was to evaluate the performance of existing Dutch syllabification algorithms across different datasets and to apply novel deep learning techniques to improve accuracy. While previous research exclusively used CELEX dictionary words to test algorithm performance \cite{Trogkanis2010,Bouma2003,Bartlett2008}, the inclusion of loanword and pseudoword datasets here shows that algorithm performance may not be consistent across different datasets. Although some algorithms achieve high accuracy under certain conditions, they may struggle when applied to other datasets. Specifically, the knowledge-based algorithm by Brandt Corstius, which performed worst on two datasets (dictionary words and loanwords), showed the highest word accuracy on the third (pseudowords). Additionally, metrics such as hyphen accuracy may not fully reflect overall performance at the word accuracy level.

The second research question explored whether adding phonetic information could enhance performance. The inclusion of phonetic information in a deep-learning model improved Dutch syllabification accuracy, resulting in a word accuracy rate of 99.65\%, a 0.10\% increase compared to the orthographic-only deep-learning model. Beyond improving accuracy, an analysis of the words where the combined model outperformed the orthographic model reveals how phonetic information contributes to accuracy. For example, Appendix~\ref{appendix:improvedwords} illustrates that Dutch contains several unique character combinations with context-dependent pronunciations. In particular, consonant sequences such as \textit{ng} can be divided in different ways depending on context, and the combined model performed better than the orthographic-only model when \textit{ng} corresponded to the phonetic notation ’N.’ Similarly, phonetic notation that differentiates between short and long vowels was often a key factor in predicting correct syllable divisions through the combined model.

Deep learning models were found to be effective at syllabifying Dutch. The orthography-based model achieved a 99.55\% word accuracy, representing a 0.04\% absolute improvement and an 8.2\% error reduction compared to the best performance reported in the literature for Dutch (99.51\%, Trogkanis and Elkan 2010). Similarly, a model combining orthographic and phonetic information achieved 99.65\% word accuracy, a 0.14\% absolute improvement and a 28.6\% error reduction compared to the best performance by \citeasnoun{Trogkanis2010}. An interesting finding concerns how deep learning models process syllabification. Internal analysis showed an error overlap of only 30\% between the deep learning and CRF models at the word level. This suggests that the deep learning model may be processing the input dataset in a different way, causing errors on different words.

In modern deep-learning architectures, transformer models have been shown to outperform BiLSTM-based models on linguistic tasks \cite{Ahmed2023}. Regarding syllabification, early exploratory testing with different deep-learning models in the current study showed that an approach using a BiLSTM setup was the most effective. This aligns with previous research comparing the performance of BiLSTM/CNN/CRF and transformer structures on Romanian syllabification, which found that a BiLSTM/CRF setup outperformed a transformer model on the same task \cite{Corlatescu2022}. There are several reasons why BiLSTM-based models may be particularly effective for syllabification. First, syllabification largely depends on local textual relationships to identify syllable boundaries. Therefore, the connection between more distant characters in a word may provide limited additional advantage compared to more localized processing. Similarly, word analysis typically involves a limited number of tokens (a maximum word length of 34 letters was used here for Dutch), which may benefit from being processed through a BiLSTM rather than a transformer. Finally, the relatively small size of the training datasets used may be insufficient for a transformer model to fully capture the nuances needed to make complex deductions about patterns.

Despite the potential of deep learning approaches, some limitations remain. First, the pseudoword dataset used here is small because a larger official pseudoword dataset is not available for Dutch. Second, the effectiveness of the combined models on the pseudoword and loanword datasets could not be tested due to the lack of available phonetic notations. Third, some Dutch algorithms from the literature were excluded from the comparison because their source code or documentation was unavailable \cite{Bouma2003,Bartlett2008}. Future research could explore modern deep learning approaches incorporating phonetic information for other orthographic processing tasks. Additionally, the combined and orthographic deep learning models developed in this study could be applied to other languages, such as English and German.

\newpage


\bibliographystyle{clin} 
\bibliography{CLIN_syllab.bib}

\newpage
\appendix

\section{Sample Words}\label{appendix:samplewords}

An overview of words contained within the three different datasets used. The first column features a sample of words present within the CELEX Dictionary Words dataset, which served as a training and testing set. The second dataset concerned a compilation of loanwords (second column). The third dataset was comprised of pseudowords (third column).

\begin{table}[h!]
\centering
\begin{tabular*}{\textwidth}{@{\extracolsep{\fill}} p{0.32\textwidth} p{0.32\textwidth} p{0.32\textwidth} @{}}
\toprule
\textbf{Dictionary Words} & \textbf{Loanwords} & \textbf{Pseudowords}  \\ \midrule
abonneren         & mesjogge                & frijk                     \\
vignet  & internet        & millen       \\
keten        & videoscherm              & sjobel                  \\
ontleer     & ciao          & zwieten          \\
dier     & yakuza          & wachteniek \\
bezorgd     & telewerken          & sprieuw   \\
welkom     & freaken          & eemde    \\
vervoeren     & cashen          & spekkeraai   \\
tijger     & fitness          & grek   \\
adoptiekind     & lawntennis          & kneurem   
\\\bottomrule
\end{tabular*}
\caption{Sample words from the three different datasets used.}\label{tab:samplewordsdatasets} 
\end{table}

\newpage

\section{Words Combined Model Improved On}\label{appendix:improvedwords}

The following table features a subsample of the words on which the combined model improved. The first two columns show the orthographic input and phonetic input the models used. The third column shows the erroneous solution the orthographic-only model output as a prediction. The last column shows the correct prediction made by the combined model, which improved on the solution provided by the orthographic only model.

In many cases, the combined model improved upon the orthographic model by relying on phonetic information that differentiated between pronunciations. For example, in \textit{geinlijn}, the vowel combination \textit{ei} was uniquely coded as \textit{K} in the phonetic notation. The orthographic-only model incorrectly marked \textit{gein} as containing a syllable break, likely because \textit{ge} is a common prefix in many Dutch words (e.g., in \textit{ge-in-ves-teerd} or \textit{ge-ind}). Likewise, the phonetic notation allows for distinction between long and short vowels. In \textit{leptosoom}, the phonetic model coded the first \textit{o} character as being a long vowel, allowing the combined model to leverage this information and correctly output \textit{lep-to-soom} over the incorrect \textit{lep-tos-oom} predicted by the orthographic model (the latter assumed the first \textit{o} to be a short vowel).

\begin{table}[!h]
\captionsetup{justification=raggedright, singlelinecheck=false} 
\centering

\begin{tabular*}{\textwidth}{@{\extracolsep{\fill}} l l l l p{0.2\textwidth} @{}}
\toprule

\multicolumn{2}{c}{\textbf{Input}} & 
\multicolumn{2}{c}{\textbf{Model}} & 
\textbf{Translation}\\

\cmidrule(r){1-2} 
\cmidrule(r){3-4} 

\textbf{Orthographic} & 
\textbf{Phonetic} & 
\textbf{Orthographic Only} & 
\textbf{Combined}
& \\ 

\midrule
suede & sy'w)d@ & sue-de & su-e-de & suede \\
leptosoom & lEpto'som & lep-tos-oom & lep-to-soom & leptosome \\
aasgieren & 'asxir@ & aas-gi-e-ren & aas-gie-ren & vultures \\
hostesses & 'hOst@s@s & hos-tes-ses & hos-tess-es & hostesses \\
care & 'k)r & ca-re & care & care \\
geinlijn & 'xKnlKn & ge-in-lijn & gein-lijn & prank call line \\
liane & li'jan@ & li-ane & li-a-ne & liane (plant) \\
anglofobie & ANGlofo'bi & ang-lo-fo-bie & an-glo-fo-bie & anglophobia \\
rizoom & 'ri'zom & riz-oom & ri-zoom & rhizome \\

\bottomrule
\end{tabular*}
\caption{Comparison between output from the orthographic model (third column) and the output from the combined model (fourth column) that predicted correct syllable-boundaries through use of phonetic information (second column).}\label{tab:datasets} 
\end{table}

\end{document}